\documentclass[11pt,a4paper]{article}
\usepackage[hyperref]{emnlp2018}
\usepackage{times}
\usepackage{latexsym}
\usepackage{amsmath}
\usepackage{url}

\usepackage{amssymb}
\usepackage{bm}
\usepackage{multirow}
\usepackage{url}
\usepackage{subfig}
\usepackage{graphicx}

\graphicspath{{pdf/}{./}}
\DeclareGraphicsExtensions{.pdf,.jpeg,.png,.jpg}

\aclfinalcopy 
\def\aclpaperid{2146} 

\title{Exploiting Contextual Information via Dynamic Memory Network for Event Detection}

\author{Shaobo Liu$^{\dag\S}$, Rui Cheng$^{\dag\S}$, Xiaoming Yu$^{\dag}$, Xueqi Cheng$^{\dag}$\\
    $^\dag$CAS Key Laboratory of Network Data Science and Technology, \\
    Institute of Computing Technology, Chinese Academy of Science (CAS); \\
    $^\S$School of Computer and Control Engineering, The University of CAS\\
  {\tt \{liushaobo, chengrui\}@software.ict.ac.cn; }\\
    {\tt \{yuxiaoming, cxq\}@ict.ac.cn}\\}
 
\date{}

\begin{document}
\maketitle
\begin{abstract}
	The task of event detection involves identifying and categorizing event triggers.
	Contextual information has been shown effective on the task.
	However, existing methods which utilize contextual information only process the context once. We argue that the context can be better exploited by processing the context multiple times, allowing the model to perform complex reasoning and to generate better context representation,
	thus improving the overall performance.
	Meanwhile, dynamic memory network (DMN) has demonstrated promising capability in capturing contextual
	information and has been applied successfully to various tasks.
	In light of the multi-hop mechanism of the DMN to model the context, we propose the trigger detection dynamic memory network (TD-DMN)
	to tackle the event detection problem. We performed a five-fold cross-validation on the ACE-2005 dataset and experimental results show that the multi-hop mechanism does improve the performance and the proposed model achieves best $F_1$ score compared to the state-of-the-art methods.
\end{abstract}

\section{Introduction}
According to ACE (Automatic Content Extraction) event extraction program, an event is identified by a word
or a phrase called event trigger which most represents that event. For example, in the sentence
``No major \emph{explosion} we are aware of'', an event trigger detection model is able to identify the word
``\emph{explosion}'' as the event trigger word and further categorize it as an \emph{Attack} event.
The ACE-2005 dataset also includes annotations for event arguments, which are a set of words or phrases that describe
the event. However, in this work, we do not tackle the event argument classification and focus on event trigger
detection.

The difficulty of the event trigger detection task lies in the complicated interaction between the event trigger candidate
and its context. For instance, given a sentence at the end of a passage:

\textbf{they are going over there to do a mission they believe in and as we said, 250 \emph{left} yesterday.}

It's hard to directly classify the trigger word ``\emph{left}'' as an ``\emph{End-Position}'' event or a
``\emph{Transport}'' event because we are not certain about what the number ``250'' and the pronoun ``they''
are referring to.
But if we see the sentence:

\textbf{we are seeing these soldiers head out.}

which is several sentences away from the former one, we now know the ``250'' and ``they'' refer to ``the soldiers'',
and from the clue ``these soldiers head out'' we are more confident to classify the trigger word ``\emph{left}''
as the ``\emph{Transport}'' event.

From the above, we can see that the event trigger detection task involves complex reasoning across the given context.
Exisiting methods \cite{liu2017exploiting,chen2015event, li2013joint,nguyen2016joint,venugopal2014relieving} mainly exploited sentence-level
features and \cite{liao2010using, zhao2018document} proposed document-level models to utilize the context.

The methods mentioned above either not directly utilize the context or only process the context once
while classifying an event trigger candidate. We argue that processing the context multiple times with later steps re-evaluating the context
with information acquired from the previous steps improves the model performance. Such a mechanism allows the model to perform complicated
reasoning across the context. As in the example, we are more confident to classify ``\emph{left}'' as a ``\emph{Tranport}'' event if
we know ``250'' and ``they'' refer to ``soldiers'' in previous steps.

We utilize the dynamic memory network (DMN) \cite{xiong2016dynamic,kumar2016ask} to capture the contextual information of the given trigger word.
It contains four modules: the input module for encoding reference text where the answer clues reside, the memory module for storing knowledge acquired from previous steps, the question module for encoding the questions, and the answer module for generating answers given the output from memory and question modules.

\begin{figure}[!t]
	\centering
	\includegraphics[width=\columnwidth]{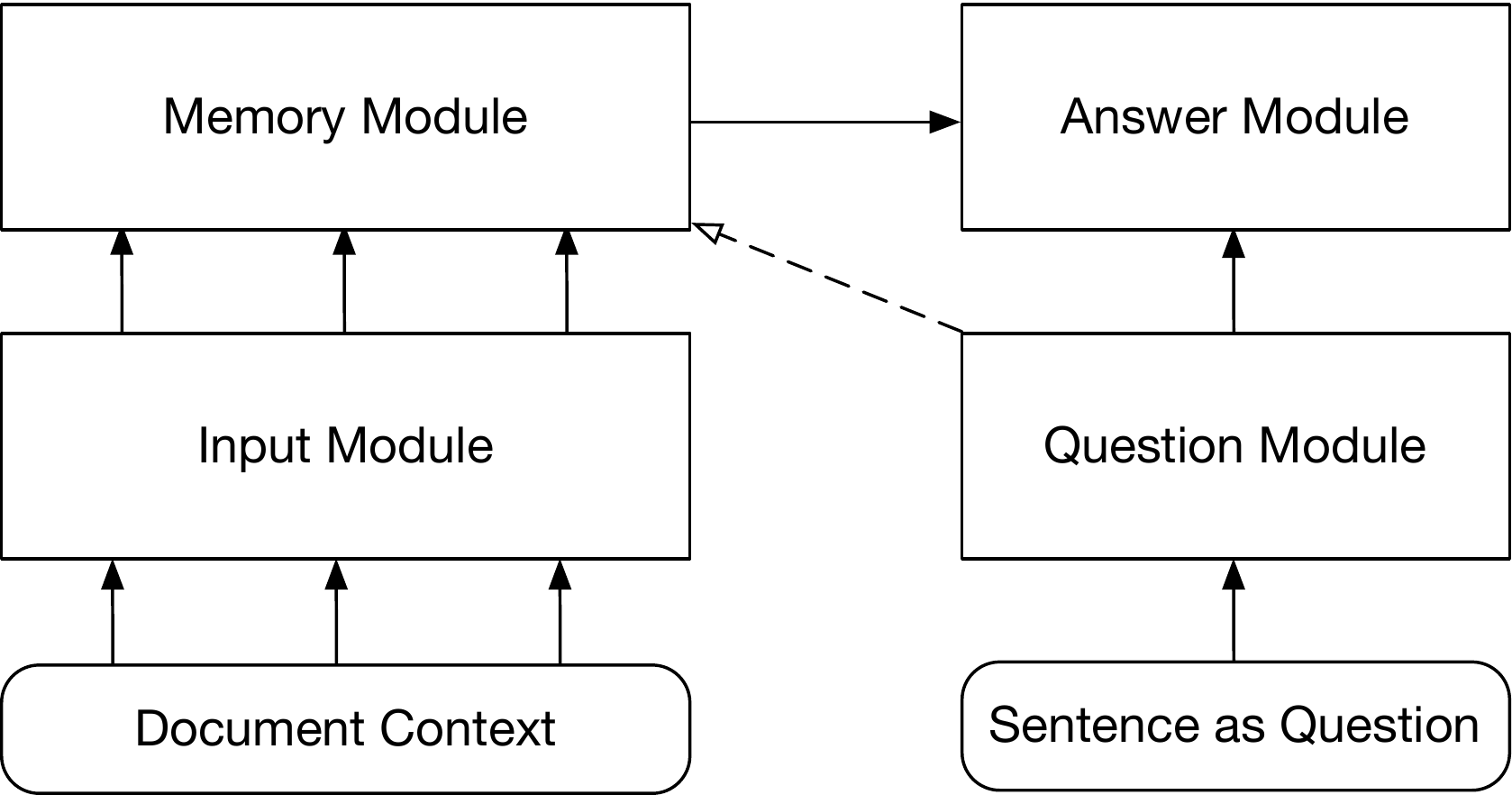}\\
	\caption{Overview of the TD-DMN architecture}
	\label{fig:overview}
	\vspace{-10pt}
\end{figure}

DMN is proposed for the question answering task, however, the event trigger detection problem does not have an explicit question.
The original DMN handles such case by initializing the question vector produced by the question module with a zero or a bias vector, while we argue that each sentence in the document could be deemed as a question. We propose the trigger detection dynamic memory network (TD-DMN) to incorporate this intuition,
the question module of TD-DMN treats each sentence in the document as implicitly asking a question ``What are the event types for the words
in the sentence given the document context''. The high-level architecture of the TD-DMN model is illustrated in Figure~\ref{fig:overview}.

We compared our results with two models: DMCNN\cite{chen2015event} and DEEB-RNN\cite{zhao2018document} through 5-fold cross-validation on the ACE-2005 dataset. Our model achieves best $F_1$ score and experimental results further show that processing the context multiple times and adding implicit questions do improve the model performance. The code of our model is available online.\footnote{https://github.com/AveryLiu/TD-DMN}。

\section{The Proposed Approach}

\begin{figure*}[!t]
	\centering
	\includegraphics[width=\textwidth]{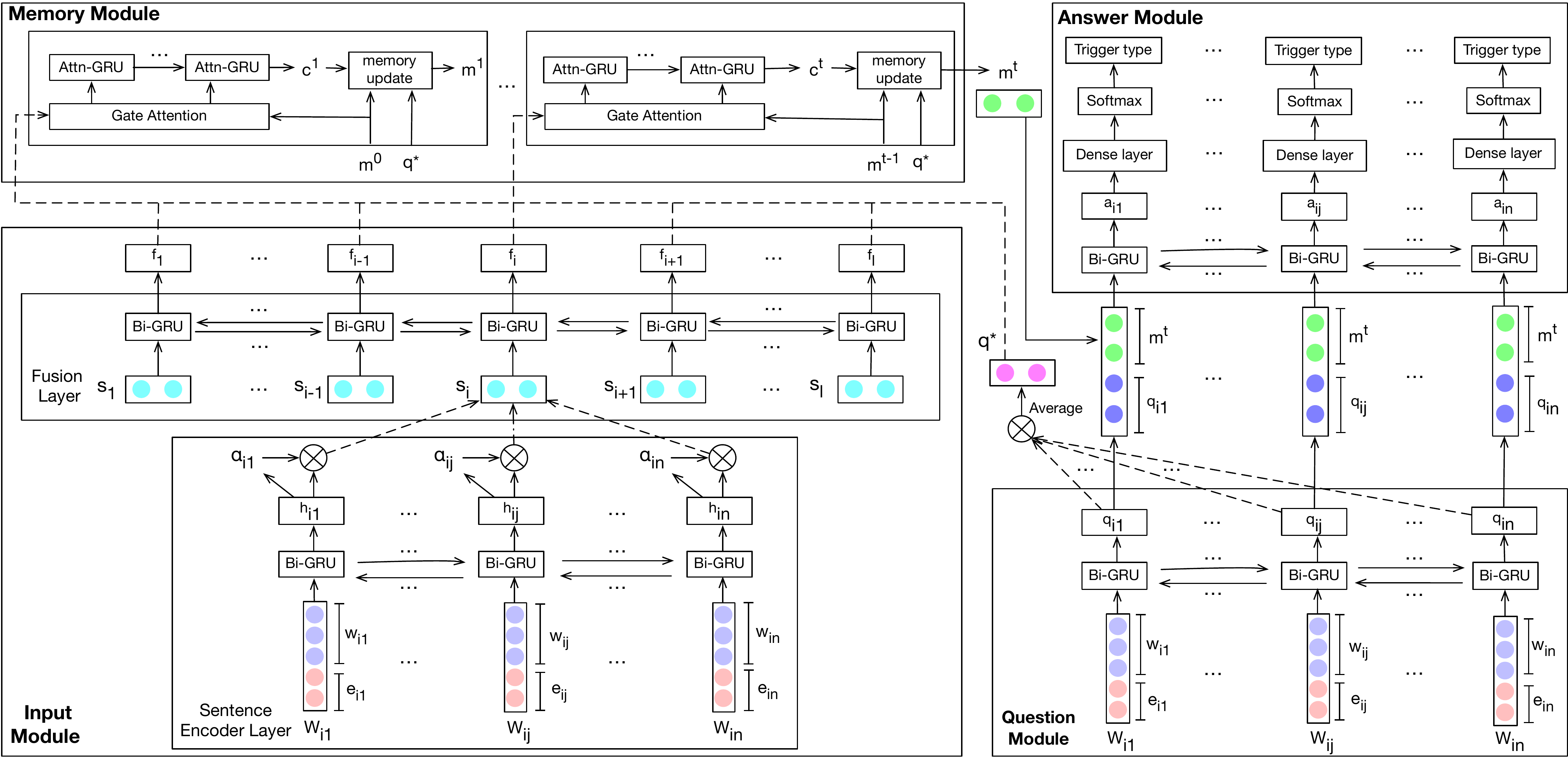}\\
	\caption{The detailed architecture of the TD-DMN model. The figure depicts a simplified case where a single document $d$ with $l$ sentences is the input
		to the input module and a sentence $s_i$ of $d$ with $n$ words is the input to the question module. The input module encodes document $d$ into a fact matrix $\{\bm{F}|\bm{f_1}, \dots, \bm{f_l}\}$. The question module encodes sentence $s_i$ into the question vector $\bm{q^*}$. The memory module initializes $\bm{m_0}$ with $\bm{q^*}$
		and iteratively processes for $t$ times, at each time $k$ it produces a memory vector $\bm{m_k}$ using fact matrix $\bm{F}$, question vector $\bm{q^*}$ and
		previous memory state $\bm{m_{k-1}}$. The answer module outputs the predicted trigger type for each word in $s_i$ using the concatenated tensor of the hidden states of the question module and the last memory state $\bm{m_t}$. }
	\label{fig:detailed}
	\vspace{-12pt}
\end{figure*}

We model the event trigger detection task as a multi-class classification problem following existing work. In the rest of this section, we describe four different modules of the TD-DMN separately along with how data is propagated through these modules. For simplicity, we discuss a single document case. The detailed architecture of our model is illustrated in Figure~\ref{fig:detailed}.

\subsection{Input Module}
The input module further contains two layers: the sentence encoder layer and the fusion layer. The sentence encoder layer encodes each sentence into a vector
independently, while the fusion layer gives these encoded vectors a chance to exchange information between sentences.

\textbf{Sentence encoder layer} Given document $d$ with $l$ sentences $(s_1, \dots, s_l)$, let $s_i$ denotes the $i$-th sentence in $d$ with $n$ words $(w_{i1}, \dots, w_{in})$. For the $j$-th word $w_{ij}$ in $s_i$, we concatenate its word embedding \bm{$w_{ij}$} with its entity type embedding\footnote{The ACE-2005 includes entity type (including type ``NA'' for none-entity) annotations for each word, the entity type embedding is a
	vector associated with each entity type.} \bm{$e_{ij}$} to form the vector \bm{$W_{ij}$} as the input to the sentence encoder Bi-GRU\cite{DBLP:journals/corr/ChoMGBSB14} of size $H_s$. We obtain the hidden state $\bm{h_{ij}}$ by merging the forward and backward hidden states from the Bi-GRU:
\begin{equation}
	\bm{h_{ij}}=\overrightarrow{{\rm{GRU}}_{s}}(\bm{W_{ij}}) + \overleftarrow{{\rm{GRU}}_{s}}(\bm{W_{ij}})
\end{equation}
where $+$ denotes element-wise addition.

We feed $\bm{h_{ij}}$ into a two-layer perceptron to generate the unnormalized attention scalar $u_{ij}$:
\begin{equation}
	u_{ij}=\tanh(\bm{h_{ij}}\cdot W_{s_1}) \cdot W_{s_2}
\end{equation}
where $W_{s_1}$ and $W_{s_2}$ are weight parameters of the perceptron and we omitted bias terms. $u_{ij}$ is then normalized to obtain scalar attention weight $\alpha_{ij}$:
\begin{equation}
	\alpha_{ij} = \frac{\exp(u_{ij})}{\sum_{k=1}^{n}\exp(u_{ik})}
\end{equation}

The sentence representation $\bm{s_i}$ is obtained by:
\begin{equation}
	{\bm{s_i}} = \sum_{j=1}^{n}\alpha_{ij}{\bm{h_{ij}}}
\end{equation}

\textbf{Fusion layer} The fusion layer processes the encoded sentences and outputs fact vectors which contain exchanged information among sentences. Let $\bm{s_i}$ denotes the $i$-th sentence representation obtained from the sentence encoder layer. We generate fact vector $\bm{f_i}$ by merging the forward and backward states from fusion GRU:
\begin{equation}
	\bm{f_i} = \overrightarrow{{\rm{GRU}}_{f}}(\bm{s_{i}}) + \overleftarrow{\rm{GRU}}_{f}(\bm{s_{i}})
\end{equation}

Let $H_f$ denotes the hidden size of the fusion GRU, we concatenate fact vectors $\bm{f_1}$ to $\bm{f_l}$ to obtain the matrix $\bm{F}$ of size $l$ by $H_f$, where the $i$-th row in $\bm{F}$ stores the $i$-th fact vector $\bm{f_i}$.

\subsection{Question Module}
The question module treats each sentence $s$ in $d$ as implicitly asking a question: What are the event types for each word in the sentence $s$ given the document $d$ as context? For simplicity, we only discuss the single sentence case. Iteratively processing  from $s_1$ to $s_l$ will give us all encoded questions in document $d$.

Let $\bm{W_{ij}}$ be the vector representation of $j$-th word in $s_i$, the question GRU generates hidden state $\bm{q_{ij}}$ by:
\begin{equation}
	\bm{q_{ij}} = \overrightarrow{{\rm{GRU}}}_{q}(\bm{W_{ij}}) + \overleftarrow{{\rm{GRU}}}_{q}(\bm{W_{ij}})
\end{equation}

The question vector $\bm{q^*}$ is obtained by averaging all hidden states of the question GRU:
\begin{equation}
	\bm{q^*} = \frac{1}{n}\sum_{j=1}^{n}\bm{q_{ij}}
\end{equation}

Let $H_q$ denotes the hidden size of the question GRU, $\bm{q^*}$ is a vector of size $H_q$, the intuition here is to obtain a vector that represents the
question sentence. $\bm{q^*}$ will be used for the memory module.

\begin{table*}[t!]
	\small
	\begin{center}
		\setlength\tabcolsep{4pt}
		\renewcommand{\arraystretch}{1.15}
		\begin{tabular}{|c|c|c|c|c|c|c|c|c|c|c|c|c|c|c|c|c|}
			\hline
			\multirow{2}{*}{Methods} &\multicolumn{3}{|c|}{Fold 1} & \multicolumn{3}{|c|}{Fold 2} & \multicolumn{3}{|c|}{Fold 3} & \multicolumn{3}{|c|}{Fold 4}&
			\multicolumn{3}{|c|}{Fold 5} & Avg\\ \cline{2-17}
			             & $P$  & $R$  & $F_1$         & $P$  & $R$  & $F_1$         & $P$  & $R$  & $F_1$         & $P$  & $R$  & $F_1$         & $P$  & $R$  & $F_1$         & $F_1$         \\ \hline
			DMCNN        & 67.6 & 60.5 & 63.9          & 62.6 & 63.1 & 62.9          & 68.9 & 62.1 & 65.3          & 68.9 & 65.0 & 66.9          & 66.0 & 65.5 & 65.8          & 64.9          \\
			DEEB-RNN     & 64.9 & 64.1 & 64.5          & 63.4 & 64.7 & \textbf{64.0} & 66.1 & 64.3 & 65.2          & 66.0 & 67.3 & 66.6          & 65.5 & 67.2 & 66.3          & 65.3          \\ \hline
			TD-DMN 1-hop & 67.3 & 62.1 & 64.6          & 65.4 & 61.7 & 63.5          & 72.0 & 60.0 & 65.4          & 66.6 & 68.0 & \textbf{67.3} & 68.3 & 65.0 & 66.6          & 65.5          \\
			TD-DMN 2-hop & 69.2 & 61.0 & 64.8          & 64.6 & 63.4 & \textbf{64.0} & 64.3 & 66.4 & 65.3          & 68.7 & 65.9 & \textbf{67.3} & 68.5 & 65.7 & \textbf{67.1} & 65.7          \\
			TD-DMN 3-hop & 66.3 & 63.7 & 64.9          & 66.9 & 60.6 & 63.6\         & 68.3 & 64.0 & 66.1          & 67.9 & 66.3 & 67.1          & 70.2 & 64.3 & \textbf{67.1} & \textbf{65.8} \\
			TD-DMN 4-hop & 66.7 & 63.4 & \textbf{65.0} & 61.4 & 65.5 & 63.4          & 66.4 & 66.0 & \textbf{66.2} & 64.7 & 69.1 & 66.8          & 70.0 & 63.4 & 66.5          & 65.6          \\ \hline
		\end{tabular}
	\end{center}
	\vspace{-10pt}
	\caption{5-fold cross-validation results on the ACE-2005 dataset. The results are rounded to a single digit. The $F_1$ of the last column are calculated
		by averaging $F_1$ scores of all folds.}
	\label{result-table}
	\vspace{-10pt}
\end{table*}

\subsection{Memory Module}
The memory module has three components: the attention gate, the attentional GRU\cite{,xiong2016dynamic} and the memory update gate.
The attention gate determines how much the memory module should attend to each fact given the facts $\bm{F}$, the question $\bm{q^*}$,
and the acquired knowledge stored in the memory vector $\bm{m_{t-1}}$ from the previous step.

The three inputs are transformed by:
\begin{equation}
	\label{eq:mu}
	\resizebox{.88\columnwidth}{!}{
	$\bm{u} = [\bm{F}*\bm{q^*}; |\bm{F}-\bm{q^*}|; \bm{F}*\bm{m_{t-1}}; |\bm{F}-\bm{m_{t-1}}|] $
	}
\end{equation}
where $;$ is concatenation. $*$, $-$ and $|.|$ are element-wise product, subtraction  and absolute value respectively.
$\bm{F}$ is a matrix of size $(m, H_f)$, while $\bm{q^*}$ and $\bm{m_{t-1}}$ are vectors of size $(1, H_q)$ and $(1, H_m)$, where $H_m$ is the output size of the memory update gate.
To allow element-wise operation, $H_f, H_q$ and $H_m$ are set to a same value $H$. Meanwhile, $\bm{q^*}$ and $\bm{m_{t-1}}$ are broadcasted to the size of $(m, H)$. In equation~\ref{eq:mu}, the first two terms measure the similarity and difference between facts and the question. The last two terms have the same functionality for facts and the last memory state.

Let $\bm{\beta}$ of size $l$ denotes the generated attention vector. The $i$-th element in $\bm{\beta}$ is the attention weight for fact $\bm{f_i}$. $\bm{\beta}$ is obtained by transforming $\bm{u}$ using a two-layer perceptron:
\begin{equation}
	\bm{\beta} = {\rm{softmax}}({\rm{tanh}}(\bm{u} \cdot W_{m_1}) \cdot W_{m_2})
\end{equation}
where $W_{m_1}$ and $W_{m_2}$ are parameters of the perceptron and we omitted bias terms.

The attentional GRU takes facts $\bm{F}$, fact attention $\bm{\beta}$ as input and produces context vector $\bm{c}$ of size $H_c$. At each time step $t$, the attentional GRU picks the $\bm{f_t}$ as input and uses $\beta_t$ as its update gate weight. For space limitation, we refer reader to \cite{,xiong2016dynamic} for the detailed computation.

The memory update gate outputs the updated memory $\bm{m_t}$ using question $\bm{q^*}$, previous memory state $\bm{m_{t-1}}$ and context $\bm{c}$:
\begin{equation}
	\bm{m_t} = {\rm{relu}} ([\bm{q^*};\bm{m_{t-1}}; \bm{c}]\cdot W_u)
\end{equation}
where $W_u$ is the parameter of the linear layer.

The memory module could be iterated several times with a new $\bm{\beta}$ generated for each time. This allows the model to attend to different parts of the facts in different iterations, which enables the model to perform complicated reasoning across sentences. The memory module produces $\bm{m_t}$ as the output at the last iteration.

\subsection{Answer Module}
Answer module predicts the event type for each word in a sentence. For each question GRU hidden state $\bm{q_{ij}}$, the answer module concatenates it with the memory vector $\bm{m_t}$ as the input to the answer GRU with hidden size $H_a$. The answer GRU outputs $\bm{a_{ij}}$ by merging its forward and backward hidden states. The fully connected dense layer then transforms $\bm{a_{ij}}$ to the size of the number of event labels $O$ and the softmax layer is applied to output the probability vector $\bm{p_{ij}}$. The $k$-th element in $\bm{p_{ij}}$ is the probability for the word $w_{ij}$ being the $k$-th event type. Let $y_{ij}$ be the true event type label for word $w_{ij}$. Assuming all sentences are padded to the same length $n$, the cross-entropy loss for the single document $d$ is applied as:
\begin{equation}
	J(\bm{y}, \bm{p}) = -\sum_{i=1}^{l}{\sum_{j=1}^{n} { \sum_{k=1}^{O}{\rm{I}}( y_{ij}= k) {\rm{log}} \, p_{ij}^{(k)}} }
\end{equation}
where $I(\cdot)$ is the indicator function.

\section{Experiments}
\subsection{Dataset and Experimental Setup}
\subsubsection*{Dateset}
Different from prior work, we performed a 5-fold cross-validation on the ACE 2005 dataset. We partitioned 599 files into 5 parts. The file names of each fold can be found online\footnote{https://github.com/AveryLiu/TD-DMN/data/splits}. We chose a different fold each time as the testing set and used the remaining four folds as the training set.

\subsubsection*{Baselines}
We compared our model with two other models: DMCNN~\cite{chen2015event} and DEEB-RNN~\cite{zhao2018document}. DMCNN is a sentence-level event detection model which enhances traditional convolutional networks with dynamic multiple pooling mechanism customized for the task. The DEEB-RNN is a state-of-the-art document-level event detection model which firstly generate a document embedding and then use it to aid the event detection task.

\subsubsection*{Evaluation}
We report precision, recall and $F_1$ score of each fold along with the averaged $F_1$ score of all folds.
We evaluated all the candidate trigger words in each testing set. A candidate trigger word is correctly classified if its event subtype and offsets match its human annotated label.

\subsubsection*{Implementation Details}
To avoid overfitting, we fixed the word embedding and added a 1 by 1 convolution after the embedding layer
to serve as a way of fine tuning but with a much smaller number of parameters. We removed punctuations, stop words and sentences which have length less equal than 2. We used the Stanford
corenlp toolkit\cite{manning-EtAl:2014:P14-5} to separate sentences. We down-sampled negative samples to ease the unbalanced classes problem.

The setting of the hyperparameters is the same for different hops of the TD-DMN model.
We set $H$, $H_s$, $H_c$, and $H_a$ to 300, the entity type embedding size to 50, $W_{s_1}$ to 300 by 600, $W_{s_2}$ to 600 by 1, $W_{m_1}$ to 1200 by 600, $W_{m_2}$ to 600 by 1, $W_u$ to 900 by 300, the batch size\footnote{In each batch, there are 10 documents.} to 10. We set the down-sampling ratio to $9.5$ and we used Adam optimizer \cite{kingma2014adam} with weight decay set to $1e^{-5}$. We set the dropout\cite{srivastava2014dropout} rate before the answer GRU to $0.4$ and we set all other dropout rates to $0.2$.
We used the pre-trained word embedding from \cite{le2014distributed}.

\subsection{Results on the ACE 2005 Corpus}
The performance of each model is listed in table~\ref{result-table}. The first observation is that models using document context drastically outperform the model that only focuses on the sentence level feature, which indicates document context is helpful in event detection task. The second observation is that increasing number of hops improves model performance, this further implies that processing the context for multiple times does better exploit the context. The model may have exhausted the context and started to overfit, causing the performance drop at the fourth hop.

The performance of reference models is much lower than that reported in their original papers. Possible reasons are that we partitioned the dataset randomly, while the testing set of the original partition mainly contains similar types of documents and we performed a five-fold cross-validation.

\subsection{The Impact of the Question Module}
To reveal the effects of the question module, we ran the model in two different settings. In the first setting, we initialized the memory vector $\bm{m_0}$ and question vector $\bm{q^*}$ with a zero vector, while in the second setting, we ran the model untouched. The results are listed in the table~\ref{question-impact-table}. The two models perform comparably under the 1-hop setting, this implies that the model is unable to distinguish the initialization values of the question vector well in the 1-hop setting. For higher number of hops, the untouched model outperforms the modified one. This indicates that with a higher number of memory iterations, the question vector $\bm{q^*}$ helps the model to better exploit the context information. We still observe the increase and drop pattern of the $F_1$ for the untouched model. However, such a pattern is not obvious with empty questions. This implies that we are unable to have a steady gain without the question module in this specific task.
\begin{table}[t!]
	\begin{center}
		\begin{tabular}{|l|l|l|}
			\hline Methods & $F_1$          & $F_1^*$ \\ \hline
			TD-DMN 1-hop   & 65.48          & 65.52   \\
			TD-DMN 2-hop   & 65.69          & 65.46   \\
			TD-DMN 3-hop   & \textbf{65.78} & 65.51   \\
			TD-DMN 4-hop   & 65.57          & 65.40   \\
			\hline
		\end{tabular}
	\end{center}
	\caption{\label{question-impact-table} The impact of the question module, $F_1^*$ indicates results with empty questions.}
	\vspace{-12pt}
\end{table}
\section{Future Work}
In this work, we explored the TD-DMN architecture to exploit document context. Extending the model to include wider contexts across several similar documents may also be of interest. The detected event trigger information can be incorporated into question
module when extending the TD-DMN to the argument classification problem. Other tasks with document context but without explicit questions may also benefit from this work.
\section{Conclusion}
In this paper, we proposed the TD-DMN model which utilizes the multi-hop mechanism of the dynamic memory network to better capture the contextual information for the event trigger detection task. We cast the event trigger detection as a question answering problem. We carried five-fold cross-validation experiments on the ACE-2005 dataset and results show that such multi-hop mechanism does improve the model performance and we achieved the best $F_1$ score compared to the state-of-the-art models.

\section*{Acknowledgments}
We thank Prof. Huawei Shen for providing mentorship in the rebuttal phase. We thank Jinhua Gao for discussion on the paper presentation. We thank Yixing Fan and Xiaomin Zhuang for providing advice regarding hyper-parameter tuning. We thank Yue Zhao for the initial discussion on event extraction. We thank Yaojie Lu for providing preprocessing scripts and the results of DMCNN. We thank anonymous reviewers for their advice. The first author personally thank Wei Qi for being supportive when he was about to give up.
\newpage
\bibliography{emnlp2018}
\bibliographystyle{acl_natbib_nourl}

\end{document}